\documentclass[sigconf]{acmart}
\usepackage{setspace}
\usepackage{microtype}
\usepackage{multirow}
\usepackage{booktabs} 
\usepackage{amsfonts}
\usepackage{amssymb}
\usepackage{graphicx}
\usepackage{float}
\usepackage{siunitx}
\usepackage{tikz} 
\usepackage{pgfplots}
\usepackage{mathtools}
\usepackage{amsmath}
\usepackage{amssymb}
\usepackage{subcaption}
\usepackage{makecell}
\usepackage{float}
\usepackage{xurl}
\usepackage{pifont}
\pgfplotsset{compat=newest} 
\usepgfplotslibrary{units} 
\usepackage{hyperref}
\usepackage{booktabs} 
\usepackage{colortbl} 
\usepackage{xcolor} 
\usepackage{xfrac}
\usepackage{enumitem}
\usepackage{booktabs} 
\usepackage{algorithm}
\usepackage{algpseudocode}
\usepackage{amsthm}

\theoremstyle{definition}

\usepackage{amssymb}
\pgfplotsset{compat=newest} 
\usepgfplotslibrary{units} 
\usepackage{listings}
\usepackage{color}

\usepackage[normalem]{ulem}
\newcommand{\pev}[1]{{\color{black} #1}}

\definecolor{mygreen}{rgb}{0,0.6,0}
\definecolor{mygray}{rgb}{0.5,0.5,0.5}
\definecolor{mymauve}{rgb}{0.58,0,0.82}
\lstdefinestyle{customc}{
	belowcaptionskip=1\baselineskip,
	breaklines=true,
	xleftmargin=\parindent,
	language=C,
	showstringspaces=false,
	basicstyle=\footnotesize\ttfamily,
	keywordstyle=\bfseries\color{green!40!black},
	commentstyle=\itshape\color{purple!40!black},
	identifierstyle=\color{blue},
	stringstyle=\color{orange},
}
\lstdefinestyle{customasm}{
	belowcaptionskip=1\baselineskip,
	breaklines=true,
	xleftmargin=\parindent,
	language=[x86masm]Assembler,
	basicstyle=\footnotesize\ttfamily,
	commentstyle=\itshape\color{purple!40!black},
}
\usepackage{subcaption}
\usepackage{amssymb}
\usepackage{bm}

\AtBeginDocument{%
  \providecommand\BibTeX{{%
    \normalfont B\kern-0.5em{\scshape i\kern-0.25em b}\kern-0.8em\TeX}}}

\setcopyright{none}
\renewcommand\footnotetextcopyrightpermission[1]{} 
\begin{document}

\title{Fine-Grained Static Detection of Obfuscation Transforms Using Ensemble-Learning and Semantic Reasoning}

\author{Ramtine Tofighi-Shirazi}
\affiliation{
	\institution{Univ. Grenoble Alpes, CNRS, Institut Fourier}
	\institution{Trusted Labs, Thales Group}
	\streetaddress{6 rue de la Verrerire}
	\city{Meudon}
	\state{France}
}
\email{ramtine.tofighishirazi@thalesgroup.com}

\author{Irina M\u{a}riuca As\u{a}voae}
\affiliation{
	\institution{Trusted Labs, Thales Group}
	\streetaddress{6 rue de la Verrerire}
	\city{Meudon}
	\state{France}
}
\email{irina-mariuca.asavoae@thalesgroup.com}

\author{Philippe Elbaz-Vincent}
\affiliation{
	\institution{Univ. Grenoble Alpes, CNRS, Institut Fourier}
	\streetaddress{100 Rue des Mathematiques}
	\city{F-38000 Grenoble}
	\state{France}
}
\email{philippe.elbaz-vincent@univ-grenoble-alpes.fr}

\renewcommand{\shortauthors}{Ramtine Tofighi-Shirazi et al.}

\begin{abstract}

 The ability to efficiently detect the software protections used is at a prime to facilitate the selection and application of adequate deobfuscation techniques. 
  We present a novel approach that combines semantic reasoning techniques with ensemble learning classification for the purpose of providing a static detection framework for obfuscation transformations.
 By contrast to existing work, we provide a methodology that can detect multiple layers of obfuscation, without depending on knowledge of the underlying functionality of the training-set used.
   We also extend our work to detect constructions of obfuscation transformations, thus providing a fine-grained methodology. 
 To that end, we provide several studies for the best practices of the use of machine learning techniques for a scalable and efficient model.
 According to our experimental results and evaluations on obfuscators such as \texttt{Tigress} and \texttt{OLLVM}, our models have up to 91\% accuracy on state-of-the-art obfuscation transformations.
 Our overall accuracies for their constructions are up to 100\%.
\end{abstract}

\keywords{machine learning, ensemble learning, deobfuscation, obfuscation, reverse engineering, symbolic execution}

\maketitle

\section{Introduction}

Code obfuscation is a widely used software protection technique to mitigate the risks of reverse-engineering.
It aims at protecting intellectual property by hiding the logic and data of a code.
The use of code obfuscation transformations depends on the sensitivity of the application.
Its applications are mainly digital right management, software licensing code or white-box cryptography, among others.
Malicious codes also use extensively code obfuscation to hide their intent, evade detection and hinder analyses.

In order to properly evaluate obfuscation transformations, or to efficiently analyze malwares, many deobfuscation techniques have emerged.
Their goal is to remove the protection layers applied on the code.
The deobfuscation process can be seen as different strategies such as reverting, simplifying, or gathering information about the obfuscated code.
In this paper we mainly focus on information gathering, particularly the static detection of obfuscation transformations.
We also study an extension to the transformations constructions, namely the different methods employed for a specific obfuscation transformation to be achieved (\textit{e.g.} dispatch-methods for control-flow flattening or code virtualization). 
This approach is previously known as \textit{metadata recovery attacks}~\cite{DBLP:conf/acsac/SalemB16}. 

State-of-the-art deobfuscation techniques are often specific to obfuscation transformations. 
For example, the work of Udupa \textit{et al.}~\cite{DBLP:conf/wcre/UdupaDM05} targets control-flow transformations, whereas others~\cite{DBLP:conf/ccs/XWW15, DBLP:conf/amast/PredaMBG06, DBLP:conf/sp/BardinDM17, DBLP:conf/acsac/Tofighi-Shirazi18} aim at removing opaque predicates.
Generic deobfuscation techniques, however, make no assumption about the applied protections~\cite{DBLP:conf/sp/YadegariJWD15, DBLP:conf/dimva/SalwanBP18}. 
These techniques are based on dynamic symbolic execution and may lack in code coverage and scalability.

Though obfuscation transformations are semantic-preserving, they may introduce side effects to the code~\cite{Ctaxonomy}. 
Each transformations has its own construction methodology, thus specific patterns.
Recent works try to tackle the detection of software protections using machine learning or deep learning techniques.
Ugarte-Pedrero \textit{et al.}~\cite{DBLP:conf/nss/Ugarte-PedreroSBGE11} propose a semi-supervised learning approach in order to classify packed and unpacked binaries.
Sun \textit{et al.}~\cite{DBLP:conf/acisp/SunVBY10}, and more recently Biondi \textit{et al.}~\cite{DBLP:journals/compsec/BiondiEGLNV19}, aim at detecting and identifying packers using machine learning techniques.
Tofighi-Shirazi \textit{et al.}~\cite{tofighishirazi:hal-02269192} propose a deobfuscation methodology for invariant opaque predicates based on machine learning techniques.

From the variety of obfuscation techniques, as well as deobfuscation methodologies, the ability to efficiently detect the software protections used is at a prime.
To that end, the recent work of Salem \textit{et al.}~\cite{DBLP:conf/acsac/SalemB16} focuses on the detection of obfuscation transformations.
Their goal is to facilitate the selection and application of adequate deobfuscation techniques. 
To the best of our knowledge, their work is the first to tackle code obfuscation detection using machine learning. 
However, their methodology is also prone to some limitations as explained next.

\paragraph{Current limitations}
Existing detection technique for code obfuscation~\cite{DBLP:conf/acsac/SalemB16} based on machine learning techniques comes with the following limitations:
\begin{enumerate}[noitemsep]
	\item \textit{Code dependency}: machine learning and syntax-reasoning used for the detection of obfuscation transformations can lead to code dependency. 
	Namely, the trained model becomes dependent to the analyzed code used in the training set, thus lowering its accuracy.
	\item \textit{Multi-class problem}: the methodology used relies on multi-class problems for classification. 
	Namely, they consider that one binary cannot be obfuscated with more than one obfuscation transformation. 
	However, transformations can be combined, thus the necessity to be able to detect the several applied layers.
	\item \textit{Granularity}: the detection technique has a high-level of granularity. 
	They may detect an obfuscation transformation, but they do not focus on their constructions.
	The latter is of importance in order to decide which analysis to apply on obfuscated code.
	Many transformations constructions are made to prevent existing deobfuscation techniques.
\end{enumerate}

\paragraph{Motivation}
\begin{figure}[h!]
	\centering
	\includegraphics[scale=0.28]{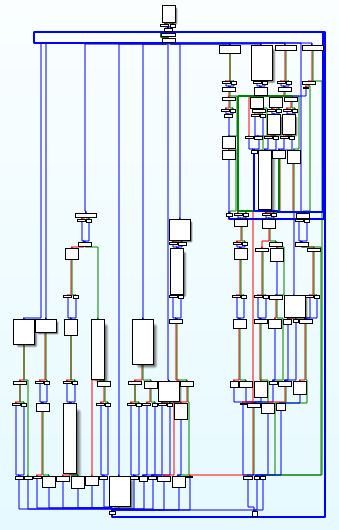}
	\caption{Control-flow graph of a quick-sort function obfuscated using several \texttt{Tigress} transformations.}
	\label{fig:running_example}	
\end{figure}
When applying obfuscation transformations for software protections, stealth is sometimes not desired.
Many applications aim for dissuasion in order to prevent reverse-engineering.
In any case, the goal of our methodology is to provide a static and automated framework to help reverse-engineers.
By detecting obfuscation transformations, and more specifically their constructions, an analyst will gain an important amount of time.
The selection of the deobfuscation process to apply requires such knowledge beforehand.
A motivating example is illustrated in Figure \ref{fig:running_example}.
It represents the obfuscated control-flow graph of a quick-sort function.
Based on the previously introduced problems, our goal is to answer the following questions:
\begin{itemize}[noitemsep]
	\item\textit{Complexity:} can we detect all applied layers of obfuscation transformation?
	\item\textit{Granularity:} can we detect the constructions of applied obfuscation transformations?
	\item\textit{Efficiency:} can we create accurate and generic enough models for unknown data?
\end{itemize}
As previously discussed in~~\cite{DBLP:conf/acsac/SalemB16}, metadata recovery attacks are usually manual tasks, therefore a potential bottleneck in the reverse engineering process.
Our methodology, which could be plugged-in a disassembler framework, provides all applied transformation and construction and allows reverse-engineers to setup automated deobfuscation strategies.
As an example, several opaque predicates constructions prevent SMT-solver based deobfuscation techniques~\cite{DBLP:conf/dsn/XuZKTL18}.
Other recent works prevent the application of dynamic symbolic execution techniques~\cite{DBLP:conf/uss/BanescuCP17, DBLP:journals/corr/abs-1908-01549}.
Thus, knowing which transformations and constructions analysts are facing may prevent using unadapted techniques for the deobfuscation process.

\paragraph{Contributions}
In order to face the above limitations and  answer  our motivating questions, we bring the following contributions:

\begin{enumerate}[noitemsep]
	\item A novel methodology that combines semantic reasoning with ensemble learning techniques applied for a multi-label and multi-output ensemble model.
	We believe that semantic reasoning will prevent our model from code dependency limitations, and provides us with the ability to detect several combined layers of obfuscation transformations.
	\item An extension of our methodology for a fine-grained detection. 
	Based on our main approach, a second classification model is used for the detection of the transformations constructions, based on a multi-class classification model (\textit{i.e.} one unique label per instances). 
	\item Several studies and experiments that justify the constructions of our methodology. 
	We compare different machine learning approaches and techniques in order to build efficient and scalable models.
	We also evaluate our methodology against state-of-the-art obfuscators such as \texttt{Tigress}~\cite{Tigress} and \texttt{Obfuscator-LLVM}~\cite{ieeespro2015-JunodRWM} (\textit{i.e.} \texttt{OLLVM}). 
\end{enumerate}

Our paper is organized as follows. Section 2 presents the background information about code obfuscation and targeted transformations.
We also introduce related work, as well as notions of supervised machine learning.
 Section 3 describes our methodology which combines semantic reasoning with ensemble learning.
Section 4 contains our studies and experiments towards an efficient implementation of our methodology.
Section 5 illustrates our evaluations on state-of-the-art and publicly available obfuscators.
Section 6 briefly discuss the application of our methodology to setup deobfuscation strategies.
Then, we discuss our design limitations in Section 7, as well as our perspectives in Section 8.
Finally,  Section 9 presents our conclusions.
\pev{\paragraph{Limitations} While our results illustrate the interest of the methodology, evaluating the exact gain of the different
components of the approach and experimental comparison to related contributions
are left as future work.}

\section{Background}
We briefly present code obfuscation and some of the employed transformations. 
Then we introduce several notions related to supervised machine learning and \textit{metadata recovery attacks} introduced in~\cite{DBLP:conf/acsac/SalemB16}.

\subsection{Code obfuscation}
Collberg \textit{et al.}~\cite{Ctaxonomy} define code obfuscation as follows: 
\\Let $P \xrightarrow{T} P'$ be a transformation $T$ of a source program $P$ into a target program $P'$. We call $P \xrightarrow{T} P'$ an
\textit{obfuscating transformation} if $P$ and $P'$ have the same observable behavior, $P'$ is harder to analyze than $P$, and $P'$ is no more than polynomially slower than $P$.	
Consequently, the following conditions must be fulfilled for an obfuscating transformation : if $P$ fails to terminate, or terminates 
with an error condition, then $P'$ may or may not terminate; otherwise, $P'$ must terminate and produce the same output as $P$.

\subsection{Obfuscation transformations}
\label{subsec:obf}
An obfuscation transformation $T$ can be classified into different categories such as data obfuscation, static code obfuscation, and dynamic code obfuscation. 
Early techniques are given by Collberg \textit{et al.}~\cite{Ctaxonomy,Collberg:2009:SSO:1594894}.
A classification of all these obfuscations, as well as known deobfuscation methods has been provided by S. Schrittwieser \textit{et al.}~\cite{DBLP:journals/csur/SchrittwieserKK16}.
The following paragraphs present a non-exhaustive list of obfuscation transformations.

\subsubsection{Encodings}
Static data within binaries, such as strings or constant values, contain useful information for an analyst. 
Encoding, as an obfuscation transformation $T$, converts data to a different representation.
To this end, special encoding functions are employed to mitigate the need of storing the static data in clear text within the binary. 
During execution, the inverse function is used to decode the obfuscated data. 
To prevent pattern-matching attacks, the obfuscated representation must be parameterized in order to have a \textit{family} of representations. 
In other words, each representation renders different-looking obfuscated variables. 
However, they are all based on the same obfuscating algorithm.

\subsubsection{Instructions substitutions}
Each program behavior can be implemented in multiple ways \cite{Warren:2012:HD:2462741}.
In other words, instructions or sequences of instructions can be replaced with syntactically different, yet semantically equivalent code.
As an example, complex instruction substitution include the replacement of call instructions with a combination of \texttt{push} and \texttt{ret} instructions~\cite{DBLP:journals/lisp/LakhotiaBSM10}.
De Sutter \textit{et al}~\cite{DBLP:conf/icisc/SutterAGCB08} replaced infrequently used opcodes with blocks of more frequently used instructions in their work. 
This transformation reduced the total number of different opcodes used in the code and normalizes their frequency.

\subsubsection{Opaque predicates}
An opaque predicate~\cite{CollbergTL98} represents an obfuscated predicate with its outcome known at obfuscation time, but difficult to determine for a deobfuscator.
Opaque predicates are used to make static reverse-engineering more complex.
They introduce an analysis problem which is difficult to solve without running the program.
There are two types of \textit{invariant} opaque predicates and the \textit{two-ways} opaque predicates.
Collberg \textit{et al.} defined these predicates by, respectively, $P^T$, $P^F$ and $P^?$ opaque predicates.
Several works use two-ways opaque predicates constructs, either referred to as range-dividers~\cite{DBLP:conf/acsac/BanescuCGNP16}, or as correlated opaque predicates~\cite{DBLP:conf/ccs/XWW15, DBLP:conf/isw/XuMW16}.
Moreover, regardless of their output, \textit{e.g.} their type, there exists many different kinds of construction that produce the opaque predicates.

\subsubsection{Control-flow flattening}
This obfuscation transformation aims at obscuring links between basic-blocks by flattening the control-flow. 
Wang \textit{et al.}~\cite{DBLP:conf/dsn/WangHKD01} describe as \textit{chenxification} this transformation, which puts the basic-blocks of a program into a large switch-statement.
A dispatcher decides then where to jump next. 
Control-flow flattening using a central dispatcher is also described by Chow \textit{et al.}~\cite{DBLP:conf/isw/ChowGJZ01}.
A similar concept by Lynn and Debray~\cite{DBLP:conf/ccs/LinnD03} uses what is called \textit{branch functions}, which directs the control-flow to the actual target based on a call table. 
Further control-flow obfuscation constructions are described in~\cite{DBLP:conf/uss/PopovDA07, DBLP:conf/ih/SchrittwieserK11, DBLP:conf/drm/CappaertP10, LazloKiss, DBLP:journals/taco/CoppensSM13}.

\subsubsection{Code virtualization}
Code virtualization describes the concept of converting a program functionality into byte-code for a custom virtual machine interpreter that is bundled with the program~\cite{DBLP:conf/sp/KingCWVWL06,DBLP:conf/ih/GhoshHD10}. 
This obfuscation transformation can also be combined with \textit{polymorphism} by implementing custom virtual machine interpreters and payloads for each instance of the program~\cite{DBLP:conf/drm/AnckaertJV06}. 
Other work~\cite{DBLP:conf/IEEEares/VrbaHG10} proposes the combination of fine-granular encryption and code virtualization to hide the virtual machine code from analysis. 
Collberg \textit{et al.}~\cite{Ctaxonomy} describe a variant of this concept under the term \textit{table interpretation}. A similar concept by Monden \textit{et al.}~\cite{DBLP:conf/acsw/MondenMT04} uses a finite state machine-based interpreter to dynamically map between instructions and their semantics.
Thus, code virtualization proposes many constructions, as for previous transformations.

\subsubsection{Dynamic code modification}
In this technique, similar functions are obfuscated by providing a general template in memory that is patched right before its execution~\cite{Ctaxonomy}.
Static analysis techniques fail to analyze the program, as its functionality is available at runtime only. 
Other concepts of dynamic code modification~\cite{DBLP:conf/compsac/KanzakiMNM03,DBLP:conf/wisa/MadouAMDSB05} implement the idea of correcting intentionally erroneous code at runtime, right before execution.
~\\

Our goal in this paper is to evaluate our methodology against the previously presented obfuscation transformations.
Beforehand, the next section will recall some notions about supervised machine learning techniques for classification.

\subsection{Supervised machine learning}
\label{subsec:supervisedml}
\textit{Supervised machine learning}~\cite{Kotsiantis:2007:SML:1566770.1566773, DBLP:books/lib/HastieTF09} provides a dedicated methodology to produce general hypotheses from external supplied instances via a given algorithm.
From these hypotheses, predictions about future instances are possible. 
The aim of a supervised machine learning is to build a \textit{classification model} which will be used to assign \textit{labels} to unknown instances.  
In other words, let $X$ be an input (\textit{i.e.} instance) and $Y$ the output (\textit{i.e.} predicted label). 
A supervised machine learning algorithm will be used to learn the mapping function $f$ such that $Y = f(X)$.
The goal is to approximate $f$ such that for any new instance $X$ we can predict its label $Y$. 
In our case the inputs are represented by $n$-dimensional vectors of numerical features for which the extraction is described in the following paragraph.
The traditional \textit{single-label} classification associates an instance $X$ with a unique label $Y'$ from a previously \emph{known} finite set of labels $L$.
This approach is then considered a \textit{binary} classification problem if $|L| = 2$, or a \textit{multi-class} classification problem if $|L| > 2$.
Other approaches exist, such as the \textit{multi-label} classification. In this case, an instance $X$ is associated with a \textbf{set} of labels $S_{Y'} \subset L$.
Moreover, if the model is based on a mapping function $f$ that can return a set of multiple labels, we have a \textit{multi-output} classification model.
In our work, we use all these classification problems as described in Section \ref{sec:methodology}.

\subsubsection{Feature extraction}
Instances of a machine learning model are usually derived from what is called \textit{raw data}, \textit{i.e.} the data samples we want to classify or predict. 
These data samples cannot be directly given to a classification model and need to be processed beforehand. 
This processing step is called \textit{feature extraction}~\cite{Guyon:2006:FEF:1208773} and consists in combining the raw data variables into numerical features.
It allows to effectively reduce the amount of data that must be processed, while accurately describing the original dataset of raw data.
In our case, raw data are text documents (\textit{e.g.} disassembly code, symbolic execution state, etc.).
Therefore, one practical use of feature extraction consists in extracting the \textit{words} (\textit{i.e.} the features) and classify them by frequency of use (\textit{i.e.} weights). 
Different approaches exist for understanding what a word is and to compute its weight.
\emph{In this paper we use the \textit{bag of words} approach~\cite{Manning:2008:IIR:1394399}, which identifies terms with words using \textit{term frequency}, in order to extract the features for our model}. It is an efficient and simple approach which fits adequately our semantic reasoning approach.
 
\subsubsection{Classification algorithms}
\label{subsec:classification_algo}
The choice of which specific learning algorithm to use is a critical step.
Many classification algorithms exist~\cite{James:1985:CA:7557}, each of them having different mapping functions. 
Classification is a common application of machine learning. As such, there are many metrics that can be used to measure and evaluate our models.
In order to compute these metrics, \textit{$k$-Fold Cross-Validation}~\cite{DBLP:conf/ijcai/Kohavi95} is a frequently used technique. 
The definition of $k$-fold cross-validation consists in reserving a particular set of samples on which the model does not train. 
The limited set of samples allows to estimate how the model is expected to perform on data not used during the training phase. 
The parameter $k$ refers to the number of groups that a given dataset of samples is split into, in order to calculate the mean of our models \textit{accuracy} as well 
as the \textit{F1-score} based on the value of $k$.
While the accuracy of the model represents the ratio of correctly predicted labels to the total of labels, F1-score takes both false positives and negatives into account.
In our experimentations and evaluations, the accuracies and F1-scores are calculated using \textit{k-fold cross-validation}, with $k=10$ for a better generalization of our model to unknown instances.

Another application of cross-validation, introduced in \cite{DBLP:conf/acsac/SalemB16}, consists in a functionality-based folding. 
In other words, the learning set and training set are divided based on the functionality of the samples from which the raw data are generated. 
The goal of such evaluation methodology is to measure if the model is dependent to the underlying code functionality, independently of the obfuscation transformation applied. The next paragraph introduce furthermore the work of Salem \textit{et al.}~\cite{DBLP:conf/acsac/SalemB16}, known as \textit{metadata recovery attack}. 

\subsection{Related work: Metadata recovery attack}
Salem \textit{et al.}~\cite{DBLP:conf/acsac/SalemB16} introduce the use of machine learning techniques to evaluate the stealth of obfuscation transformations throughout their detection (known as \textit{metadata recovery attack}). 
Their primary hypothesis is that machine learning techniques are capable of implementing these attacks by classifying obfuscated programs according to the transformations applied.
Their experiments are based on two learning algorithms, namely Naive Bayes~\cite{Friedman:1997:BNC:274158.274161} and Decision trees~\cite{Rokach:2014:DMD:2755359}.
Their raw data are based on static disassembly or dynamic instruction traces, either stripped or not. 
Thus, we refer to such raw data generation as \textit{syntax-reasoning}.
The evaluation of their models is made with two classification techniques. 
The first one is a traditional $k$-fold cross validation, with $k=10$. 
The second one is more fine-tuned since it discriminates the training and test dataset on program functionality. 
In other words, the test dataset is excluded of any raw data that have been used in the training dataset, based on the functionality they implement. 
Such process is also repeated 10 times, to calculate the average accuracy for each fold.
Their results are promising, showing up to 100\% of accuracy for obfuscation transformations detection with decision trees, on dynamic traces. 
However, these results are obtained with the conventional cross-validation, whereas the second classification mode provides lower results (up to 61\% of accuracy) with decision trees.
This indicates that their model is dependent of the functionality implemented in their raw data.
Moreover, their work is not implemented yet to cover several layers of obfuscation transformations, as it can be the case in most obfuscated programs.
In our work, we also used both cross-validation approach to compare our results with their work. This gives an brief idea about the advantages of semantic reasoning over syntax-based approaches.

~\\
Our goal in this paper is to combine semantic reasoning and more advanced machine learning classification techniques \pev{in order to improve the accuracy}.
We want to have a static analysis tool, based on symbolic execution, in order to have a model that does not depend on the functionality of the program.
The models are used to detect several layers of obfuscation transformations, thus having a multi-label and multi-output classification problem.
Then, we extend our detection not only to the obfuscation transformations but also to their constructions.
To this end, in the next section, we present our approach and methodology.

\begin{figure*}[ht!]
	\centering
	\includegraphics[width=\textwidth]{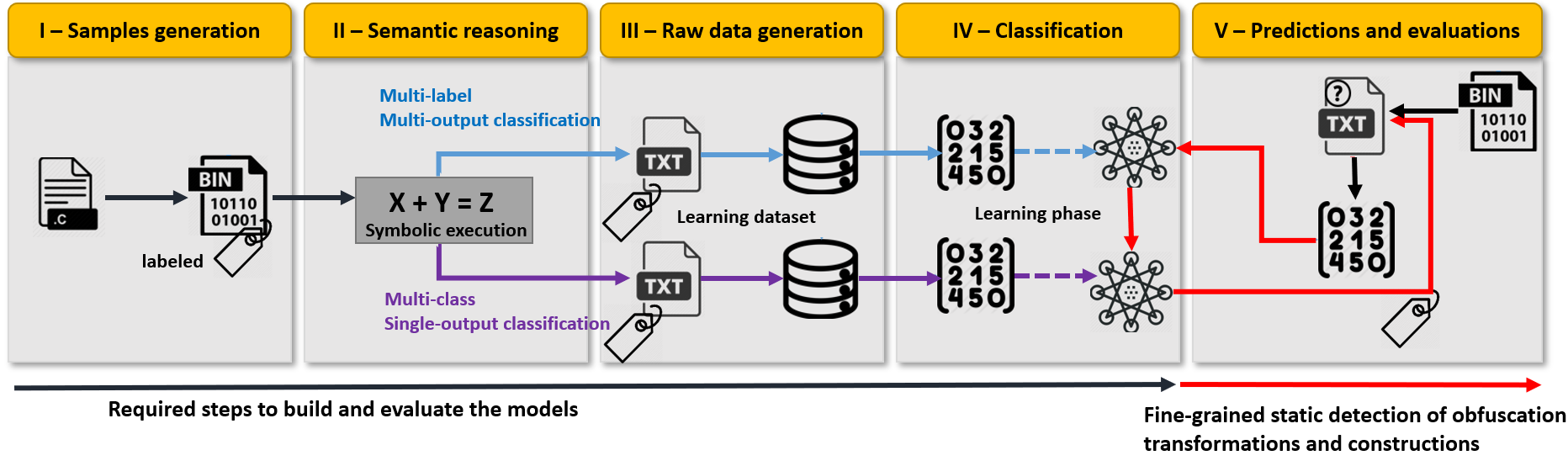}
	\vspace{-20pt}
	\caption{Design steps for fine-grained static detection of obfuscation transformations and constructions.}
	\label{fig:design}	
\end{figure*}

\section{Methodology}
\label{sec:methodology}
In this section we present our methodology composed of several steps, as illustrated in Figure \ref{fig:design}.
\textbf{I.} In order to create our models, we  generate obfuscated as well as clean samples.
This generation  is done using publicly available obfuscators, specifically \texttt{Tigress} and \texttt{OLLVM}.
\textbf{II.} We employ then semantic reasoning via symbolic execution\footnote{In our work we consider semantic retrieval only. We are not interested in generating inputs for program exploration.} to extract our raw data, from the generated samples.
This step is presented in Section \ref{subsec:symbexec}.
\textbf{III.} We create two different datasets for two different kinds of classifications.
Using labeled raw data, we build our datasets for the detection of obfuscation transforms, including several combinations.
Another dataset is made for the detection of specific constructions related to the transformations.
These steps are introduced in Section \ref{subsec:class}.
\textbf{IV.} The previous datasets are used to train our models.
In order to select the most relevant approach and learning algorithms, several studies and experiments are provided in Section \ref{subsec:studies}.
\textbf{V.} The final step consists in their evaluation and their application on unknown instances, as presented in Section \ref{sec:evals}.

\subsection{Semantic reasoning}
\label{subsec:symbexec}
Static symbolic execution is a binary analysis technique that captures the semantics (\textit{i.e.} logic) of a program. 
An interpreter is used to trace the program, while assuming symbolic values for inputs rather than obtaining concrete values as a normal execution would. 
A symbolic state $\mathbb{S}$ is built and consists in a set of symbolic expressions $S$ for each variables (\textit{i.e.} registers, memory, flags, etc.).
Several techniques exist for symbolic execution~\cite{DBLP:journals/csur/BaldoniCDDF18}.
\pev{In order to avoid path explosions in static symbolic execution, we use an intra-procedural and bloc-centric approach, as summarized next.}

\subsubsection{Bloc-centric intra-procedural symbolic execution}
We use semantic reasoning for the generation of our \textit{raw data}.
The symbolic representation helps to efficiently detect obfuscation transformations and constructions.
Raw data refers to the representation of data samples, containing noisy features, which need to be processed in order to extract the informative characteristics to train the models.
For the detection of obfuscation transformations, we choose to work on disassembled functions of binary code. 
On these functions, we apply static symbolic execution to retrieve their semantic representation.
In our work we use disassembled functions to collect the symbolic expressions from the code, as illustrated in Algorithm \ref{algo:sem}.
First, the semantic reasoning part of our methodology is given a disassembled function $F$ as input.
For the learning phase of our methodology, $F$ needs to be labeled.
In other words, we need to know which transformations are applied in order to properly train our model.
However, in order to use our methodology as a static and automated detection framework, $F$ does not require to be labeled once the models are trained.
\begin{algorithm}
	\caption{semantic reasoning for raw-data generation}
	\label{algo:sem}
	\begin{algorithmic}[1]
		\Procedure{semantic reasoning}{$F$: a disassembled function}
		\State Initialize a dictionary $L$ 
		\For{each basic block $B$ in $F$}
		\State $I_{B} \gets $ \texttt{getInstructions}($B$)
		\State $IR_{B} \gets $\texttt{getIntermediateLanguage}($I_{B}$)
		\State $S_{B} \gets $\texttt{symbolicExecution}($IR_{B}$)
		\State $NS_{B} \gets $\texttt{normalizeSemantics}($S_{B}$)
		\State $L[B] \gets NS_{B}$
		\EndFor
		\State \texttt{textFile} = \texttt{generateRawData}($L$, $F$)
		\State \Return \texttt{textFile}
		\EndProcedure
	\end{algorithmic}
\end{algorithm}
Based on $F$, we iterate over each basic block $B$.
We then collect the instructions of $B$, denoted by $I_B$, with the function $getInstructions()$.
$I_B$ is translated into an intermediate language, denoted by $IR_B$, using \texttt{getIntermediateLanguage()}. 
Finally, $IR_B$ is being used for the bloc-centric symbolic execution function \texttt{symbolicExecution()}.
The latter will return the symbolic state $S_B$, in order words, expressions of each modified variables in a static single assignment form, based on the intermediate representation $IR_{B}$ previously used.
The generated semantics $S_B$ is then normalized using \texttt{normalizeSemantics()} function.
Finally, the normalized semantics $NS_B$ is added to the dictionary $L$ containing all normalized semantics for each processed basic block $B$.
The content of $L$ will be used to generate our raw data as text file.
Our normalization step has the crucial role of making the model scale to unknown data. 
Next, Section \ref{subsec:raw_data} describes this step, along with the content of our raw data.

\subsection{Semantic-based raw data}
\label{subsec:raw_data}
Intermediate representations often use concrete values within their generated expressions. 
This causes raw data to depend on addresses that are specific to some binaries and prevents our models to scale on unknown data. 
Some intermediate representations also use identifiers in order to express modified registers or memory areas. 
\begin{lstlisting}[belowskip=-0.8 \baselineskip,basicstyle=\tiny,language=Python,caption={Symbolic state using Miasm2 intermediate language}\label{lst:m2}]
ExprMem(ExprOp('+', ExprId('RSP_init', size=64), ExprInt(0xffffffffffffffd0, 64)), size=64) = ExprId('RDI_init', size=64)
ExprId('af', size=1) = ExprSlice(ExprOp('^', ExprOp('+', ExprId('RSP_init', size=64), ExprInt(0xffffffffffffffc8, 64)), ExprOp('+', ExprId('RSP_init', size=64), ExprInt(0xfffffffffffffff8, 64)), ExprInt(0x30, 64)), 4, 5)
ExprId('RBP', size=64) = ExprOp('+', ExprId('RSP_init', size=64), ExprInt(0xfffffffffffffff8, 64))
ExprMem(ExprOp('+', ExprId('RSP_init', size=64), ExprInt(0xffffffffffffffc8, 64)), size=32) = ExprSlice(ExprId('RDX_init', size=64), 0, 32)
ExprId('pf', size=1) = ExprOp('parity', ExprOp('&', ExprMem(ExprInt(0x606078, 64), size=64), ExprInt(0xff, 64)))
ExprId('RAX', size=64) = ExprMem(ExprInt(0x606078, 64), size=64)
ExprId('IRDst', size=64) = ExprCond(ExprMem(ExprInt(0x606078, 64), size=64), ExprInt(0x40064b, 64), ExprInt(0x400644, 64))
ExprId('zf', size=1) = ExprCond(ExprMem(ExprInt(0x606078, 64), size=64), ExprInt(0x0, 1), ExprInt(0x1, 1))
ExprMem(ExprOp('+', ExprId('RSP_init', size=64), ExprInt(0xfffffffffffffff8, 64)), size=64) = ExprId('RBP_init', size=64)
ExprId('of', size=1) = ExprInt(0x0, 1)
ExprId('nf', size=1) = ExprSlice(ExprMem(ExprInt(0x606078, 64), size=64), 63, 64)
ExprId('cf', size=1) = ExprInt(0x0, 1)
ExprId('RSP', size=64) = ExprOp('+', ExprId('RSP_init', size=64), ExprInt(0xffffffffffffffc8, 64))
ExprId('RIP', size=64) = ExprInt(0x400650, 64)
ExprId('IRDst', size=64) = ExprInt(0x400650, 64)
\end{lstlisting}
This notation may further affect the scalability of our trained models.
For the purpose of having a model that can scale to unknown data we use a normalization phase.
The normalization consists in replacing all identifiers and concrete values by symbols, and non-alphanumerical characters by alphanumerical words. 
This is a necessary step for a complete features extraction phase that sometimes excludes non-alphanumerical characters when working on text-based raw data. 
In our methodology, we generate the raw data using the \texttt{Miasm2}~\cite{Miasm} intermediate language. 
This language is part of the symbolic execution engine that we use for the implementation of our methodology as \texttt{IDA Pro} plug-in.
Additionally, the normalized \texttt{Miasm2} intermediate language has also been successful for the application of machine learning techniques in order to deobfuscate opaque predicates~\cite{tofighishirazi:hal-02269192}.
Listing 1 illustrates the symbolic state $S$ of the first basic-block of the function quick-sort, which is illustrated in Figure \ref{fig:running_example}. 
Note that the complete raw data will contain the symbolic states of each basic-blocks of the quick-sort function.
We can see that \texttt{Miasm2} intermediate language uses several keywords to express the semantics of the basic blocks.
For example, \texttt{ExprId} is used for registers and \texttt{ExprInt} for concrete values.
The registers and concretes values prevent our model from scaling to unknown data, thus potentially lowering our model accuracy.
This underlines the necessity to normalize the intermediate language for an efficient semantic reasoning.
Listing \ref{lst:nm2} illustrates the same basic-block symbolic state, but normalized.
\begin{lstlisting}[belowskip=-0.8 \baselineskip,basicstyle=\tiny,language=Python,caption={Symbolic state using our normalized Miasm2 intermediate language}\label{lst:nm2}]
ExprMem(ExprOp(op+, REG0, v0), size=64) = REG1 
REG2 = ExprSlice(ExprOp(op^, ExprOp(op+, REG0, v1), ExprOp(op+, REG0, v2), v3), 4, 5)
REG3 = ExprOp(op+, REG0, v2)
ExprMem(ExprOp(op+, REG0, v1), size=32) = ExprSlice(REG4, 0, 32)
REG5 = ExprOp(opparity, ExprOp(op&, ExprMem(v4 ,size=64), v5))
REG6 = ExprMem(v4, size=64)
IRDst = ExprCond(ExprMem(v4, size=64), v7, v8)
REG8 = ExprCond(ExprMem(v4, size=64), v9, v10)
ExprMem(ExprOp(op+, REG0, v2), size=64) = REG9 
REG10 = v9 
REG11 = ExprSlice(ExprMem(v4, size=64), 63, 64)
REG12 = v9 
REG13 = ExprOp(op+, REG0, v1)
REG14 = v11 
IRDst = v11 
\end{lstlisting}
Additionally, the normalization step also reduces the size of the raw data.
This helps enhancing the efficiency of learning and testing phase in terms of execution time.
The next sections will present the different machine learning techniques used in our methodology.
The purpose is to create automated and efficient models for the detection of obfuscation transformations, as well as their constructions.

\subsection{Ensemble learning}
\label{subsec:ensemble}
In machine learning, ensemble methods~\cite{Dietterich:2000:EMM:648054.743935} use multiple learning algorithms.
They are mostly used to obtain better predictive performance than could be obtained from any of the constituent learning algorithms alone \cite{DBLP:journals/air/Rokach10, DBLP:journals/corr/abs-1106-0257}. 
An \textit{ensemble}, in this case, consists of a set of individually trained classifiers whose predictions are combined when processing novel instances.
Different families of ensemble learning methods exists, e.g. Bagging~\cite{DBLP:journals/ml/Breiman96b}, Boosting~\cite{DBLP:conf/colt/Freund90, DBLP:conf/icml/FreundS96} or Stacking~\cite{DBLP:journals/ml/SmythW99}.
Since every model has its strengths and weaknesses, ensemble models combine individual models to help cope with the weaknesses of each algorithms.
In order to select the best possible predictions from our ensemble, we use a \textit{voting}~\cite{DBLP:journals/ijids/SuKG09} algorithm. 
Hence, a model is selected to make the final prediction by a simple majority vote for accuracy. 
Our work aims to study the benefits of ensemble learning approach over individual models.
Thus, \emph{we base our core methodology on voting classifiers}. However, a more in-depth studies of other approaches could provide better insights into the reasons why/if ensemble models get \pev{consistenly} better results for this task.

\subsection{Multi-label and multi-class classifications}
\label{subsec:class}
Multi-label classification methods are increasingly required by modern applications ~\cite{DBLP:conf/ismir/LiM03, DBLP:journals/pr/BoutellLSB04}.
We use multi-label with multi-output classification, in order to return all the detected obfuscation layers, specially when combined.
We also focus on multi-class classifications \pev{which play a key role in our methodoly due to the following facts:}
\begin{enumerate}[noitemsep]
	\item the detection of all the applied obfuscation transformations is a \emph{multi-label classification problem}. 
	For example, if our set of labels are the applied transformations, namely control-flow flattening and code virtualization, then one binary can have both protections. In such case, our methodology needs to return all predicted labels. We then refer to such model as a \emph{multi-output classification}.
	\item the fine-grained detection of the constructions is a \emph{multi-class classification problem}. 
	For example, if we know that control-flow flattening is applied on a code, then its constructions can only be one unique label (e.g. switch-based, ifnest-based, indirect, call-based, etc.).
\end{enumerate}
Multi-label classification methods differ from binary or multi-class approaches.
Tsoumakas \textit{et al.}~\cite{DBLP:journals/jdwm/TsoumakasK07} group multi-label classification methods into two categories: \textit{problem transformation methods} that transform the multi-label classification problem either into one or more single-label classification problems, and
\textit{algorithm adaptation methods} that extend specific learning algorithms in order to handle multi-label data directly.
In our methodology we use classifier chains~\cite{DBLP:journals/ml/ReadPHF11}, where each model is an ensemble of learning algorithm, as presented in Section \ref{subsec:ensemble}.
We also study the \textit{binary relevance} methodology~\cite{DBLP:conf/pakdd/GodboleS04} in Section \ref{subsec:studies}.
These two methodologies are briefly introduced in the following paragraphs.

\subsubsection{Problem transformation methods}
the \textit{binary relevance} method \cite{Zhang:2018:BRM:3201424.3201455} is a problem transformation technique that transforms any multi-label problem into a binary problem for each label.
Hence, it trains several classifiers, one for each class, \textit{i.e.} one per obfuscation transformations. 
The union of all classes that are predicted is taken as the multi-label output. 
Binary relevance method is popular due to its easy implementation.
However, the main drawback is that it ignores the possible correlations between labels.
\textit{Classifier chains}~\cite{Read:2011:CCM:2070617.2070629} however, as opposed to binary relevance method, take into account the labels correlations.
With this methodology we have for $n$ labels also $n$ binary classifiers $f_0, f_1, ..., f_n$ constructed.
The construction is made as a chain where a classifier $f_i$ uses the predictions of all its previous classifiers $f_j$ with $j < i$.
The chain order is randomly selected in our design.

\subsubsection{Algorithm adaptation methods}
Algorithm adaptation extends single label classification to the multi-label context.
It is usually done by changing the decision functions.
Some learning algorithms support multi-label and multi-output classification (\textit{e.g.}~\cite{Zhang:2007:MLL:1234417.1234635,Zhang:2006:MNN:1159162.1159294}), whereas others can be extended.

~\\ 
During our experiments, these two classifications approaches, and multi-label problems will be studied in Section \ref{subsec:studies}.
Our objective is to provide the best suited algorithms and techniques for an efficient and accurate model.

\section{Experiments}
\label{subsec:studies}
In this section we present first the dataset used, common with previous related work~\cite{tofighishirazi:hal-02269192, DBLP:conf/acsac/SalemB16}.
Our preliminary studies towards an efficient implementation of a fine-grained detection framework are also introduced.
All our experiments and evaluations are done on a Windows 7 laptop, using 16GB of RAM, and an Intel processor.

\subsection{Datasets}
\label{subsec:datasets}
Our experiments are made on several C code samples. 
We use the \texttt{scikit-learn} API~\cite{scikit-learn} for the implementation of the models.
The datasets contain various types of code, each of them having different functionalities in order to have models that do not fit to a specific type of program.
The used samples are listed below:
\begin{itemize}
	\item GNU core utilities (\textit{i.e.} core-utils) binaries~\cite{Coreutils} for normal predicate samples;
	\item Cryptographic binaries for obfuscated and non-obfuscated predicates~\cite{Bcon};
	\item Samples from~\cite{DBLP:conf/acsac/BanescuCGNP16} containing basic algorithms (\textit{e.g.} factorial, sorting, etc.), non-cryptographic hash functions, small programs generated by \texttt{Tigress};
	\item Samples involving the uses of structures and aliases~\cite{fragglet, thealgo}.
\end{itemize}
Our choice is motivated by the samples low ratio of dependencies and their straightforward compilation.
This makes their obfuscation possible using tools such as \texttt{Tigress} and \texttt{OLLVM} without errors during compilation.
\emph{Furthermore, all datasets used for the studies and evaluations are balanced and contain between 1000 to 5000 samples.} The obfuscation transformations applied are given in Appendix \ref{appendix:ollvm} and \ref{appendix:tigress}.
The next section will present our studies based on these datasets.

\subsection{Preliminary studies}
Our goal in this section is to provide some answers to the following questions related to our methodology:
\begin{itemize}[noitemsep]
	\item \textbf{Study 1}: when only one obfuscation transformation is applied, is a single model more effective than ensemble models for the detection?
	\item \textbf{Study 2}: when several obfuscation transformations are applied, can the model from Study 1 be applied to the multi-label and multi-output classification problems?
	\item \textbf{Study 3}: when several obfuscation transformations are applied, is a multi-label and multi-output model more efficient than one binary model for each transformation, \textit{i.e.} classifier chains?
	\item \textbf{Study 4}: for the fine-grained detection of obfuscation constructions, is a single model more efficient than ensemble models?
\end{itemize}
Our  studies and evaluations present two different types of results based on two different evaluations approaches.
One is the traditional $k$-folds cross-validation with scores in black colored font.
The other is made with the functionality-based cross-validation approach in red colored font, used in Salem \textit{et al.} related work~\cite{DBLP:conf/acsac/SalemB16}.
Besides, we use as a traditional single-model random-forest algorithm throughout all our studies.
As for the ensemble models, we combined extra-tree and random-forest learning algorithms.
These algorithms were selected because they provided the best scores in terms of accuracy. 
For simplicity, a preliminary evaluation was made between several learning algorithms~\cite{DBLP:journals/informaticaSI/Kotsiantis07} (\textit{e.g.} decision trees, $k$-nearest neighbors, support vector machines, neural network, naive Bayes, random forest, etc.).
In order to select the best ensemble models, we combined between $2$ to $6$ single models, and selected the combination that provided the best scores.

\subsubsection{Study 1:}
In this study we experiment traditional models against ensemble learning for multi-class classification problems.
Namely, each sample is assigned with a \textit{unique} label. 
Thus our model returns only one label per sample.
We experiment here if ensemble learning can be more efficient at detecting obfuscation transformation, when only one layer is applied.
Therefore, we do not combine obfuscation transformations for this study.
\begin{table}[h]
	\centering
	\resizebox{\columnwidth}{!}{
		\begin{tabular}{|c|c|c|}
			\hline 
			\rowcolor{cyan!10} \textbf{Obfuscation transformation} & \textbf{Mono-model} & \textbf{Ensemble-learning}\\
			\hline 
			\rowcolor{black!10} \textbf{Tigress transformations} & \textbf{Random-forest} & \textbf{Extra-tree \& Random-forest}\\
			\hline 
								\textbf{EncA} & 93\% / \textcolor{red}{98\%} & 95\% / \textcolor{red}{100\%}\\
			\rowcolor{black!10}\textbf{EncL} & 100\% / \textcolor{red}{97\%} & 100\% / \textcolor{red}{100\%}\\
								\textbf{EncD} & 95\% / \textcolor{red}{98\%} & 95\% / \textcolor{red}{100\%}\\
			\rowcolor{black!10}\textbf{AddO} & 100\% / \textcolor{red}{100\%} & 98\% / \textcolor{red}{100\%}\\
								\textbf{Flat} & 97\% / \textcolor{red}{100\%} & 97\% / \textcolor{red}{100\%}\\
			\rowcolor{black!10}\textbf{Virt} & 100\% / \textcolor{red}{100\%} & 100\% / \textcolor{red}{100\%}\\
								\textbf{Jit} & 100\% / \textcolor{red}{100\%} & 100\% / \textcolor{red}{100\%}\\
			\rowcolor{black!10}\textbf{clean} & 91\% / \textcolor{red}{100\%} & 91\% / \textcolor{red}{100\%}\\
					\textbf{Overall Accuracy} & 97\% / \textcolor{red}{99\%} & 97\% / \textcolor{red}{100\%}\\
			\hline 
		\end{tabular}
	}
	\caption{Multi-class accuracy and F1-scores per labels for the detection of \texttt{Tigress} obfuscation transformations (1 layer).}
	\label{tab:study1}
\end{table}
\vspace{-6mm}
Table \ref{tab:study1} illustrates our results where we see that ensemble-learning provides a similar accuracy to random-forest, up to 97\%, with traditional cross-validation. 
The illustrated F1-scores per labels, namely the obfuscation transforms, also points out that most of them are predicated similarly with both approaches.
An exception is made for arithmetic encoding, \textit{i.e. EncA}, and opaque predicates, \textit{i.e. AddO}.
With the functionality-based cross-validation approach however, the results differs more as observed in red font.
Ensemble-learning technique provides 100\% accuracy and F1-score for each classes, whereas random-forest achieves slightly lower results, with an average accuracy at 99\%. 
Due to the semantic reasoning of our methodology, the results are better with this approach when having one layer of obfuscation.
Yet, these results are not sufficient to select traditional mono-models over ensemble-learning, or the opposite way. 
Hence, the next study will experiment these two approaches for multi-label and multi-output classification. 

\subsubsection{Study 2:}
In the following study, we combine all obfuscation transformations.
The goal of our model is to correctly predict all the applied layers of obfuscation transformation.
Thus, each sample can have one or more labels.
We aim to compare the random-forest algorithm with the ensemble model based on random-forest and extra-trees for multi-label and multi-output classification.
\begin{table}[!ht]
	\centering
	\resizebox{\columnwidth}{!}{
		\begin{tabular}{|c|c|c|}
			\hline 
			\rowcolor{cyan!10} \textbf{Obfuscation transformation} & \textbf{Multi-label mono-model} & \textbf{Multi-label ensemble}\\
			\hline 
			\rowcolor{black!10} \textbf{Tigress transformations} & \textbf{Random-forest} & \textbf{Extra-tree \& Random-forest}\\
			
			\hline 
			\textbf{EncA} & 95\% / \textcolor{red}{93\%} & 96\% / \textcolor{red}{92\%}\\
			\rowcolor{black!10}\textbf{EncL} & 90\% / \textcolor{red}{78\%} & 92\% / \textcolor{red}{85\%}\\
			\textbf{EncD} & 95\% / \textcolor{red}{93\%} & 96\% / \textcolor{red}{92\%}\\
			\rowcolor{black!10}\textbf{AddO} & 96\% / \textcolor{red}{88\%} & 97\% / \textcolor{red}{88\%}\\
			\textbf{Flat} & 98\% / \textcolor{red}{97\%} & 99\% / \textcolor{red}{91\%}\\
			\rowcolor{black!10}\textbf{Virt} & 99\% / \textcolor{red}{98\%} & 99\% / \textcolor{red}{99\%}\\
			\textbf{Jit} & 100\% / \textcolor{red}{95\%} & 97\% / \textcolor{red}{92\%}\\
			\rowcolor{black!10}\textbf{clean} & 90\% / \textcolor{red}{90\%} & 91\% / \textcolor{red}{87\%}\\
			\textbf{Overall Accuracy} & 90\% / \textcolor{red}{83\%} & 92\% / \textcolor{red}{82\%}\\
			\hline 
		\end{tabular}
	}
	\caption{Multi-label accuracy and F1-scores per labels for the detection of \texttt{Tigress} obfuscation transformations (several layers).}
	\label{tab:study2}
\end{table}
\vspace{-6mm}
Our results in Table \ref{tab:study2} illustrate that traditional cross-validation provides a higher overall accuracy for ensemble learning classifier as opposed to random forest.
Our ensemble of models scores 92\% as opposed to 90\% for random-forest, with F1-scores per labels above 91\%.
The functionality-based cross-validation provides lower results, with an overall accuracy at 83\% and at 82\% for respectively random forest and ensemble models.
Still, our result indicates that both approaches can efficiently detect several layers of obfuscation transforms. 
However, we may improve our results using problem transformations methods such as classifier chains.

The next study will experiment this hypothesis. 

\subsubsection{Study 3:}
As in the second study, we combine all obfuscation transformations but we use binary classification problem for multi-label and multi-output classification using classifier chains.
\begin{table}[h]
	\centering
	\resizebox{\columnwidth}{!}{
		\begin{tabular}{|c|c|c|}
			\hline 
			\rowcolor{cyan!10} \textbf{Obfuscation transformation} & \textbf{Mono-model chain} & \textbf{Ensemble chain}\\
			\hline 
			\rowcolor{black!10} \textbf{Tigress transformations} & \textbf{Random-forest} & \textbf{Extra-tree \& Random-forest}\\
			\hline 
						   \textbf{EncA} & 95\% / \textcolor{red}{92\%} & 97\% / \textcolor{red}{90\%}\\
				\rowcolor{black!10}\textbf{EncL} & 90\% / \textcolor{red}{80\%} & 93\% / \textcolor{red}{87\%}\\
				\textbf{EncD} & 95\% / \textcolor{red}{92\%} & 97\% / \textcolor{red}{96\%}\\
				\rowcolor{black!10}\textbf{AddO} & 96\% / \textcolor{red}{92\%} & 97\% / \textcolor{red}{88\%}\\
				\textbf{Flat} & 97\% / \textcolor{red}{97\%} & 99\% / \textcolor{red}{91\%}\\
				\rowcolor{black!10}\textbf{Virt} & 99\% / \textcolor{red}{98\%} & 99\% / \textcolor{red}{99\%}\\
				\textbf{Jit} & 100\% / \textcolor{red}{90\%} & 100\% / \textcolor{red}{92\%}\\
				\rowcolor{black!10}\textbf{clean} & 88\% / \textcolor{red}{90\%} & 92\% / \textcolor{red}{90\%}\\
				\textbf{Overall Accuracy} & 90\% / \textcolor{red}{85\%} & 92\% / \textcolor{red}{90\%}\\
			\hline 
		\end{tabular}
	}
	\caption{Classifier chain accuracy and F1-scores per labels for the detection of \texttt{Tigress} obfuscation transformations (several layers).}
	\label{tab:study3}
\end{table}
\vspace{-6mm}
Our results with standard cross-validation does not different much from previous Study 2 as illustrated in Table \ref{tab:study3}.
The functionality-based cross-validation provides improved overall accuracies and F1-scores per labels.
Ensemble models used in classifier chains provide 90\% of overall accuracy, compared to random-forest used in classifier chains that score 85\% of overall accuracy.
This study led us to select ensemble-learning techniques with classifier chains in our methodology since classifier chains allow us to create an efficient and accurate model for the detection of obfuscation transformations with one or more layers.

\subsubsection{Study 4:}
For this final study, our goal is to evaluate the models for the fine-grained detection of an obfuscation transformation construction.
\begin{table}[!ht]
	\centering
	\resizebox{\columnwidth}{!}{
		\begin{tabular}{|c|c|c|}
			\hline 
			\rowcolor{cyan!10} \textbf{Code virtualization} & \textbf{Mono-model} & \textbf{Ensemble model}\\
			\hline 
			\rowcolor{black!10} \textbf{Tigress constructions} & \textbf{Random-forest} & \textbf{Extra-tree \& Random-forest}\\
			\hline 
			\textbf{linear-based} & 100\% / \textcolor{red}{99\%} & 100\% / \textcolor{red}{100\%}\\
			\rowcolor{black!10}\textbf{switch-based} & 100\% / \textcolor{red}{98\%} & 100\% / \textcolor{red}{100\%}\\
			\textbf{if-nest-based} & 100\% / \textcolor{red}{100\%} & 100\% / \textcolor{red}{100\%}\\
			\rowcolor{black!10}\textbf{Overall Accuracy} & 100\% / \textcolor{red}{99\%} & 100\% / \textcolor{red}{100\%}\\
			\hline 
		\end{tabular}
	}
	\caption{Accuracy and F1-scores per labels for the detection of Virtualized constructions.}
	\label{tab:study4}
\end{table}
\vspace{-6mm}
We use in our dataset several Virtualized samples with \texttt{Tigress} for our experiment.
\texttt{Tigress} allows the user to select different kinds of constructions, such as \textit{switch-based, ifnest-based, linear-based, interpolation-based} for example.
This experiment is equivalent to Study 1 in the sense that it is a multi-class classification problem.
Namely, each sample has a unique label and the selected model will return one unique label per instance.

Our results in Table \ref{tab:study4} show that both random-forest and ensemble models provides the same F1-scores per labels.
Their overall accuracies with standard cross-validation are also with 100\% accuracy.
With functionality-based cross-validation, ensemble models are \pev{slightly} more efficient with a 100\% accuracy as opposed to 99\% for mono-model based on random-forest.
This led us to select ensemble models in our methodology also for the classification of constructions, as it allows a fined-grained detection capability.

\section{Evaluations}
\label{sec:evals}
In this section we evaluate our models with respect to the following classification problems:
\begin{enumerate}[noitemsep]
	\item\emph{Multi-label and multi-output evaluation:} can our model, based on a classifier chain of ensemble models, efficiently and accurately detect all obfuscation transformations when one or more layers are applied?
	\item\emph{Multi-class evaluation:} once the obfuscation transformation detected, can our ensemble model efficiently and accurately detect the construction of the latter?
\end{enumerate}
We use both cross-validation evaluation schemes as detailed in Section \ref{subsec:classification_algo}.
Our evaluations are made with publicly available obfuscators, specifically \texttt{Tigress} and \texttt{OLLVM}, in order to combine obfuscation transformations from different tools.

\subsection{Transformations detection}
Our goal is to evaluate the stealth of obfuscation transformation, either applied as unique layer or combined.
We use our multi-label and multi-output model based on ensemble-models and classifier chain to detect all the transformations applied.
To measure the efficiency of our model, we used both traditional and functionality-based cross-validation as explained in Section \ref{subsec:classification_algo}.
A list of all combinations of the applied transformations used in our evaluations can be found in Appendices \ref{appendix:tigress} and \ref{appendix:ollvm}.
Additionally, command line options for \texttt{Tigress} and \texttt{OLLVM} are given in \ref{cmd:tig} and \ref{cmd:ollvm}.

\begin{table}[h]
	\centering
	\footnotesize
	\begin{tabular}{|c|c|}
		\hline 
		\rowcolor{cyan!10} \textbf{Obfuscation transformation} & \textbf{Classifier Chain}\\
		\hline 
		\rowcolor{black!10} \textbf{OLLVM} & \textbf{Ensemble model}\\
		\hline 
		\textbf{bcf} & 98\% / \textcolor{red}{98\%} \\
		\rowcolor{black!10}\textbf{fla} & 92\% / \textcolor{red}{95\%} \\
		\textbf{sub} & 82\% / \textcolor{red}{80\%} \\
		\rowcolor{black!10}\textbf{clean} & 94\% / \textcolor{red}{93\%} \\
		\textbf{Overall Accuracy} & 86\% / \textcolor{red}{89\%} \\
		\hline
		\rowcolor{black!10}\textbf{Cross-validation execution time} & 11s for 1000 samples\\
		\hline
	\end{tabular}
	
	\caption{Evaluated accuracy and F1-scores per labels for the detection combined \texttt{OLLVM} transformations.}
	\label{tab:ollvm}
\end{table}
\vspace{-6mm}

\subsubsection{OLLVM}
Our first evaluation uses \texttt{OLLVM}.
It implements transformations such as opaque predicates (\textit{i.e.} bogus control flow, \textit{bcf}), instruction substitutions (\textit{i.e. sub}) and control-flow flattening (\textit{i.e. fla}).
We built a dataset with several combinations of these transformations (\textit{c.f.} Appendix \ref{appendix:ollvm}) in order to measure the efficiency of our model.
Table \ref{tab:ollvm} shows our results.
Our model achieves an overall accuracy of 86\% with traditional cross-validation and 89\% with the functionality-based one. 
F1-scores for labels \textit{bcf}, \textit{fla}, and \textit{clean} where no transformations are applied, are over 92\% and up to 98\% for \textit{bcf}.
However, the efficiency of our model to detect \texttt{OLLVM} instructions substitutions transformations, labeled as \textit{sub}, achieves a low F1-score at 80\%.
Further evaluations indicate that \textit{sub} is often considered \textit{clean} by our model.
Thus, when combined with other transformations, \textit{sub} transformation is often undetected.

\subsubsection{Tigress}
Our second evaluation is made with the \texttt{Tigress} obfuscator.
\texttt{Tigress} can generate state-of-the-art transformations such as dynamic-code generation (\textit{i.e. Jit}), code-virtualization (\textit{i.e. Virt}), control-flow flattening (\textit{i.e. Flat}), opaque predicates (\textit{i.e. AddO}) and several encoding (\textit{i.e. Arithmetics, Literals} and \textit{Data}, respectively \textit{EncA, EncL} and \textit{EncD}), among others.
\begin{table}[h]
	\centering
\footnotesize
		\begin{tabular}{|c|c|}
			\hline 
			\rowcolor{cyan!10} \textbf{Obfuscation transformation} & \textbf{Classifier Chain}\\
			\hline 
			\rowcolor{black!10} \textbf{Tigress} & \textbf{Ensemble model}\\
			\hline 
							\textbf{EncA} & 94\% / \textcolor{red}{90\%} \\
		\rowcolor{black!10}\textbf{EncL} & 90\% / \textcolor{red}{86\%}\\
							\textbf{EncD} & 92\% / \textcolor{red}{91\%} \\
		\rowcolor{black!10}\textbf{AddO} & 95\% / \textcolor{red}{96\%} \\
							\textbf{Flat} & 96\% / \textcolor{red}{98\%} \\
		\rowcolor{black!10}\textbf{Virt} & 99\% / \textcolor{red}{100\%} \\
							\textbf{Jit} & 100\% / \textcolor{red}{100\%} \\
		\rowcolor{black!10}\textbf{clean} & 91\% / \textcolor{red}{89\%} \\
				\textbf{Overall Accuracy} & 90\% / \textcolor{red}{91\%} \\
		\rowcolor{black!10}\textbf{Cross-validation execution time} & 114s for 4000 samples\\
		\hline
	\end{tabular}

\caption{Evaluated accuracy and F1-scores per labels for the detection combined \texttt{Tigress} transformations.}
\label{tab:tigress}
\end{table}
\vspace{-6mm}
As illustrated in Table \ref{tab:tigress}, our model accuracy is up to 90\% with standard cross-validation. 
With functionality-based cross-validation, the overall accuracy is at 91\%.
F1-scores for heavy transformation such as \textit{Virt} and \textit{Jit} are up to 99\% and 100\%.
The lowest F1-score is for \textit{i.e. EncL} which is sometimes considered as a \textit{clean} sample by our model.
Regardless, our evaluation underlines the accuracy and efficiency of our methodology against \texttt{Tigress} transformations.

\subsubsection{OLLVM and Tigress}
For this evaluation we combine both \texttt{OLLVM} and \texttt{Tigress} datasets.
We aim to see if our model is able to detect common obfuscation transformations.
\begin{table}[h]
	\centering
\footnotesize
		\begin{tabular}{|c|c|}
			\hline 
			\rowcolor{cyan!10} \textbf{Obfuscation transformation} & \textbf{Classifier Chain}\\
			\hline 
			\rowcolor{black!10} \textbf{Tigress and OLLVM} & \textbf{Ensemble model}\\
			\hline 
								\textbf{EncA and sub} & 93\% / \textcolor{red}{90\%} \\
					\rowcolor{black!10}\textbf{EncL} & 88\% / \textcolor{red}{88\%}\\
										\textbf{EncD} & 90\% / \textcolor{red}{88\%} \\
			\rowcolor{black!10}\textbf{AddO and bcf} & 95\% / \textcolor{red}{95\%} \\
								\textbf{Flat and fla} & 96\% / \textcolor{red}{99\%} \\
					\rowcolor{black!10}\textbf{Virt} & 99\% / \textcolor{red}{100\%} \\
										\textbf{Jit} & 100\% / \textcolor{red}{100\%} \\
					\rowcolor{black!10}\textbf{clean} & 83\% / \textcolor{red}{80\%} \\
							\textbf{Overall Accuracy} & 88\% / \textcolor{red}{86\%} \\
			\rowcolor{black!10}\textbf{Cross-validation execution time} & 143s for 5000 samples\\
			\hline
		\end{tabular}
	
	\caption{Evaluation accuracy and F1-scores per labels for the detection of both \texttt{Tigress} and \texttt{OLLVM} transformations.}
	\label{tab:mixed}
\end{table}
\vspace{-6mm}
Table \ref{tab:mixed} shows our results. 
F1-scores for heavy transformations such as \textit{Virt}, \textit{Jit} and \textit{Flat} are high, averaging up to 100\% for \textit{Jit} as an example.
Combined test samples between obfuscators such \textit{EncA-sub}, \textit{AddO-bcf}, and \textit{Flat-fla} have high F1-scores, even when combined with other transformations.
These heavy transformations introduce important side-effects to the code, allowing an efficient and accurate detection of our model.
The ability to efficiently detect non-obfuscated samples is still low compared to the ability to detect all layers of obfuscation transformations.
In that case, our model F1-scores are up to 83\% and 80\% depending on the cross-validation approach used.
Still, our model is averaging an accuracy up to 88\% and 86\%.
These overall accuracies illustrate our model efficiency regarding the detection of obfuscation transformations, even when combined, and between the two different obfuscators.

\subsubsection{OLLVM vs. Tigress}
Our final evaluation aims to compare the accuracies of our model depending on the learning dataset used.
First, we use a learning dataset only based on \texttt{OLLVM} transforms. The model will be then evaluated against some similar obfuscation transformations generated by \texttt{Tigress}. Second, we do the opposite, namely train our model on \texttt{Tigress} samples to evaluate it on \texttt{OLLVM} raw data.
The results are displayed in Table \ref{tab:finaleval}.
\begin{table}[h]
	\centering
	\footnotesize
	\begin{tabular}{|c|c|c|}
		\hline 
		\rowcolor{cyan!10} \textbf{Training dataset} & \textbf{Testing dataset}& \textbf{Overall accuracy}\\
		\hline 
		\textbf{OLLVM} & Tigress (Flat) & 100\% / \textcolor{red}{100\%} \\
		\rowcolor{black!10}\textbf{OLLVM} & Tigress (Flat, AddO) & 68\% / \textcolor{red}{61\%}\\
		\textbf{Tigress} & OLLVM (fla) & 95\% / \textcolor{red}{92\%} \\
		\rowcolor{black!10}\textbf{Tigress} & OLLVM (all) & 82\% / \textcolor{red}{75\%} \\
		\hline 
	\end{tabular}
	
	\caption{Overall accuracies of our model using either \texttt{OLLVM} or \texttt{Tigress} learning dataset.}
	\label{tab:finaleval}
\end{table}
\vspace{-6mm}
As we can see, our model efficiently detects \texttt{Tigress} \textit{Flat} transformation when training on 1000 samples of all \texttt{OLLVM} transforms, with 100\% of accuracy. 
Results are lower when the training dataset is based on \texttt{Tigress} (4000 samples), against \texttt{OLLVM} \textit{fla} transform, with an overall accuracy up to 95\% with a standard cross-validation.
Moreover, we can observe that our model cannot efficiently detect \texttt{Tigress} opaque predicates, \textit{i.e. AddO}, when training only on \texttt{OLLVM} transforms.
The results, in that case, indicate that our model efficiently detects the \textit{Flat} transformation, but only few \textit{AddO} ones.
Finally, when our model is trained on \texttt{Tigress}, the overall accuracy is up to 82\% against all OLLVM transforms (c.f. Appendix \ref{appendix:ollvm}). This result indicates that our methodology provides some genericity.

\subsection{Constructions detection}
In this section we evaluate our model for the detection of specific obfuscation transformations constructions.
We use our multi-class model, based on ensemble-models, to provide a fine-grained detection technique.
As for previous evaluations, we use traditional and functionality-based cross-validation techniques.
\begin{table}[!ht]
	\centering
	\footnotesize
	\begin{tabular}{|c|c|}
		\hline 
		\rowcolor{cyan!10} \textbf{Control-flow flattening} &  \textbf{Ensemble model}\\
		\hline 
		\rowcolor{black!10} \textbf{Tigress and OLLVM} &  \textbf{Extra-tree \& Random-forest}\\
		\hline 
		\textbf{switch-based} & 98\% / \textcolor{red}{95\%} \\
		\rowcolor{black!10}\textbf{if-nest-based} & 98\% / \textcolor{red}{100\%} \\
		\textbf{Overall Accuracy} & 98\% / \textcolor{red}{97\%} \\
		\rowcolor{black!10}\textbf{Cross-validation execution time} & 12s for 1000 samples\\
		\hline 
	\end{tabular}
	
	\caption{Evaluation accuracy and F1-scores per class for the detection of control-flow flattening constructions.}
	\label{tab:flat}
\end{table}
\vspace{-6mm}
\subsubsection{Control-flow flattening}
As for code virtualization, control-flow flattening can also be constructed in several ways, as introduced in Section \ref{subsec:obf}.
Facing the same limitations as for code virtualization constructions, we evaluated two constructions namely \textit{switch-based} from the \texttt{Tigress} obfuscator, and \textit{ifnest-based} from \texttt{OLLVM}.
The evaluation results are in Table \ref{tab:flat}. 
Our model averages high F1-scores and accuracy, the latter being at 98\% with standard cross-validation evaluation.

\subsubsection{Opaque predicates}
Many opaque predicates constructions exists, some of them having as purpose preventing the usage of existing deobfuscation techniques based on dynamic-symbolic execution.
For the detection of their constructions, we used \texttt{Tigress}, \texttt{OLLVM} but also novel \texttt{bi-opaque} methods~\cite{DBLP:conf/dsn/XuZKTL18}.
Our results in Table \ref{tab:addo} show that our model is accurately detecting opaque predicates constructions.
F1-scores are up to 100\% with standard cross-validation. 
Bi-opaque constructions are however often un-detected when combined with other transformations.

\begin{table}[!ht]
	\centering
	\footnotesize
	\begin{tabular}{|c|c|}
		\hline 
		\rowcolor{cyan!10} \textbf{Opaque predicates}  & \textbf{Ensemble model}\\
		\hline 
		\rowcolor{black!10} \textbf{Tigress and OLLVM}  & \textbf{Extra-tree \& Random-forest}\\
		\hline 
		\textbf{Floats} & 85\% / \textcolor{red}{89\%} \\
		\rowcolor{black!10}\textbf{Symbolic-memory} & 87\% / \textcolor{red}{87\%} \\
		\textbf{Arithmetic} & 100\% / \textcolor{red}{100\%} \\
		\rowcolor{black!10}\textbf{Aliasing} & 100\% / \textcolor{red}{99\%}\\
		\textbf{Mixed-boolean and arithmetic} & 100\% /  \textcolor{red}{96\%} \\
		\rowcolor{black!10}\textbf{Overall Accuracy} & 95\% /  \textcolor{red}{93\%} \\
		\textbf{Cross-validation execution time} & 24s for 1000 samples\\
		\hline
	\end{tabular}
	
	\caption{Evaluation accuracy and F1-scores per class for the detection of opaque predicates constructions.}
	\label{tab:addo}
\end{table}
\vspace{-8mm}
Yet, the overall accuracy of our model is at 95\% and 93\% depending on the evaluation approach used.
This illustrates the efficiency of our methodology towards the detection of obfuscation transformations constructions.

\section{Limitations}
\label{sec:threats}

One threat to the validity of our results is that we only use datasets of relatively small C programs, except for the core-utils binaries used for non-obfuscated samples. 
\pev{Nevertheless, the samples used in our dataset involve}  all common programming language constructions and \pev{various} functionalities (\textit{e.g.} hash functions, sorts, cryptographic algorithms, etc.). However, our future work will include the evaluation of our methodology on other obfuscators or programs, such as malwares.
Our work shows that semantic reasoning combined with advanced machine learning present capabilities for a fine-grained detection of obfuscation transforms.

The capability of detecting \textit{unknown} transformations or constructions represents another limitation of our methodology.
If our model did not train on one specific transformation or constructions, it will not predict properly the unknown sample.
This can lead to a loss of accuracy when unknown transformations are combined.

Dynamic transformations cause limitations to our model for the \textit{static} detection of obfuscation transforms.
Despite from the fact that we are able to accurately detect some of these transformations (\textit{i.e.} \textit{Jit}, \textit{Virt}), when other obfuscation transformations are applied before them, our model is less efficient.
Moreover, other transformations such as packing, or anti-symbolic execution techniques may lower the accuracy of our model.
However, as pointed out in the next section, our methodology can scale to dynamically collected traces which allows to thwart some of these limitations.

\section{Perspectives and future work}

First, more in-depth studies of aggregation approaches used in ensemble learning must be done in order to \pev{assess} if ensemble learning are \pev{consistenly} more efficient for that task compared to mono-models.
The hard voting scheme used is a simple approach, but may not achieve the \pev{effective benefit of using} the ensemble learning approach.

\pev{As seen in~\cite{tofighishirazi:hal-02269192},  semantic reasoning and machine learning provides promising results for deobfuscation methodology.
The evaluations shown in this paper illustrate that our model does not depend on the code functionality.}  A more accurate comparison must be made as future work.


To overcome the dynamic transformations limitations, 
\pev{we can adapt our methodology} to dynamically collected instructions traces.
With a given instructions trace, we reconstruct each basic-blocks and apply our semantic reasoning approach in order to generate raw data.
This step can be done either for the learning or the evaluation phase.
Our future work consists in extending the implementation of our \pev{current} framework  and evaluating other combinations of obfuscation transformations based on dynamic traces.

Another \pev{issue} we need to consider is the application of $n$ layers of the same obfuscation transformations.
\pev{Presently}, our evaluations is done by combining several transformations, but using one time each of them. \pev{Our future study should consider the extension of our evaluations to the use of one transformation several times.}

\pev{We also plan on extending our datasets of C programs with  more complex real-world software libraries in the interest of strengthening our experiments.}

\section{Conclusions}
\label{sec:conclusion}
In this paper we presented the efficiency of semantic reasoning combined with advanced machine learning techniques.
This combination is motivated by the construction of a fine-grained detection framework of obfuscation transformations and constructions.
By extending our approach to multi-label and multi-output classification, we enhanced metadata recovery attacks to the detection of multiple layers of obfuscation transformations.
We proposed a new approach that combines a bloc-centric symbolic execution with machine learning ensemble model and classifier chains.
We used our models to evaluate the stealth of both obfuscation transformations and constructions.
Our results are promising, with overall accuracies up to 91\% for the transformations and 100\% for the constructions, \pev{showing slight improvements with respect to current mono-models machine learning.}
The use of static symbolic execution allows us to be dependent on the underlying functionality of the code samples used for the learning phase.
Our empirical studies illustrate that our choices conduct towards the implementation of an efficient and accurate evaluation framework against state of the art obfuscators.
However, there is still place for improvements with a more in-depth study of learning algorithms used and their parameters.
Our work \pev{slightly} improves metadata-recovery attacks, and paves the way towards the efficient use of advanced machine learning combined with semantic reasoning.
\begin{acks}
	We are much grateful to the referees comments and suggestions from the editorial board.
	This work is supported by the French National Research Agency in the framework of the Investissements d'Avenir program (ANR-15-IDEX-02).
\end{acks}

\appendix
\section{Tigress transformations}
\label{appendix:tigress}
We list  the combinations of obfuscation transformations used for our datasets, in their application order:
\small{AddOpaque (16 or 32 times); AddOpaque, EncodeLiterals; EncodeLiterals; AddOpaque, EncodeArithmetics; EncodeArithmetics, AddOpaque; EncodeArithmetics; AddOpaque, EncodeData; EncodeData, AddOpaque; EncodeData; AddOpaque, EncodeArithmetics, EncodeLiterals, EncodeData; EncodeData, EncodeArithmetics, EncodeLiterals, AddOpaque; AddOpaque, Flatten ; Flatten, AddOpaque; Flatten; Flatten, EncodeData, EncodeArithmetics, EncodeLiterals; Virtualize, AddOpaque; Virtualize; Virtualize, EncodeData, EncodeArithmetics, EncodeLiterals; Virtualize, Flatten; Flatten, AddOpaque, EncodeData, EncodeArithmetics, EncodeLiterals; Virtualize, AddOpaque, EncodeData, EncodeArithmetics, EncodeLiterals; Virtualize, Flatten, AddOpaque, EncodeData, EncodeArithmetics, EncodeLiterals; Jit; Jit, AddOpaque; Jit, AddOpaque, EncodeData, EncodeArithmetics, EncodeLiterals.}

\subsection{Commands options}
\label{cmd:tig}
\begin{lstlisting}[basicstyle=\tiny,language=Python,caption={Tigress commands for sample generation}\label{lst:app_listing_tigress}]
# AddOpaque options
tigress --Transform=InitEntropy --Transform=InitOpaque --InitOpaqueStructs=list,array,env --Functions=main --Transform=AddOpaque --Functions=${3} --AddOpaqueCount=${NUM} --AddOpaqueKinds=call,fake,true
# Flatten
tigress --Transform=Flatten --FlattenDispatch=switch,goto --Functions=${3}
# Virtualize
tigress --Transform=Virtualize --VirtualizeDispatch=switch,direct,ifnest,linear --Functions=${3}
# Jit
tigress -include $TIGRESS_HOME/jitter-amd64.c --Transform=Jit --Functions=${3} --JitEncoding=hard 
# EncodeLiterals
tigress --Transform=EncodeLiterals --Functions=${3} --EncodeLiteralsKinds=integer,string
# EncodeArithmetics
tigress --Transform=EncodeArithmetic --Functions=${3} --EncodeLiteralsKinds=integer
# EncodeData
tigress --Transform=EncodeData --LocalVariables=${4} --EncodeDataCodecs=poly,xor,add --Functions=${3}\end{lstlisting}

\section{OLLVM transformations}
\label{appendix:ollvm}
We list the combinations of obfuscation transformations used for our datasets, in their application order:
\small{bcf; bcf, sub; bcf, sub, fla; bcf, fla, sub; sub; sub, bcf; sub, bcf, fla
; fla
; fla, bcf
; fla, sub, bcf
; fla, bcf, sub.}

\subsection{Commands options}
\label{cmd:ollvm}
\begin{lstlisting}[basicstyle=\tiny,language=Python,caption={OLLVM commands for sample generation}\label{lst:app_listing_ollvm}]
# Bogus control-flow
clang ${1}.c -o ${1} -mllvm -bcf -mllvm -bcf_prob=50
clang ${1}.c -o ${1} -mllvm -bcf -mllvm -bcf_prob=100
# Control-flow flattening
clang ${1}.c -o ${1} -mllvm -fla
clang ${1}.c -o ${1} -mllvm -fla -mllvm -split
# Instruction substitution
clang ${1}.c -o ${1} -mllvm -sub\end{lstlisting}
\bibliography{bibliographie_these}
\bibliographystyle{plain}

\end{document}